# A Latent Dirichlet Allocation (LDA) Semantic Text Analytics Approach to Explore Topical Features in Charity Crowdfunding Campaigns

## Prathamesh Muzumdar ᵃ*, George Kurian ᵇ and Ganga Prasad Basyal ᶜ

*ᵃ The University of Texas at Arlington, USA.*
*ᵇ Eastern New Mexico University, USA.*
*ᶜ Black Hills State University, USA.*

*Original Research Article*

## ABSTRACT

Crowdfunding in the realm of the Social Web has received substantial attention, with prior research examining various aspects of campaigns, including project objectives, durations, and influential project categories for successful fundraising. These factors are crucial for entrepreneurs seeking donor support. However, the terrain of charity crowdfunding within the Social Web remains relatively unexplored, lacking comprehension of the motivations driving donations that often lack concrete reciprocation. Distinct from conventional crowdfunding that offers tangible returns, charity crowdfunding relies on intangible rewards like tax advantages, recognition posts, or advisory roles. Such details are often embedded within campaign narratives, yet, the analysis of textual content in charity crowdfunding is limited. This study introduces an inventive text analytics framework, utilizing Latent Dirichlet Allocation (LDA) to extract latent themes from textual descriptions of charity campaigns. The study has explored four different themes, two each in campaign and incentive descriptions. Campaign description's themes are focused on child and elderly health mainly the ones who are diagnosed with terminal diseases. Incentive description's themes are based on tax benefits, certificates, and appreciation posts. These themes, combined with numerical parameters, predict campaign success. The study was successful in using Random Forest Classifier to predict success of the campaign using both thematic and numerical parameters. The study distinguishes thematic categories, particularly medical need-based charity and general causes, based on project and incentive descriptions. In conclusion, this research bridges the gap by showcasing topic modelling utility in uncharted charity crowdfunding domains.

## 1. INTRODUCTION

In recent times, charity crowdsourcing has emerged as a prevalent strategy for fundraising. It enables beneficiaries to gather financial support from a broad audience interested in causes aligning with their preferences. These contributors donate varying amounts [1]. Despite their multitude, people typically contribute small sums across numerous campaigns featured on charity crowdfunding platforms [2]. Presently, there exists competition among individual beneficiaries and nonprofit entities for funds from a sizable yet limited pool of donors [3]. Charity campaigners leverage their prior campaign know-how to enhance the likelihood of achieving their fundraising objectives [4]. Lately, incentive-driven crowdfunding platforms have significantly aided campaign organizers in securing project funds by offering concrete or intangible rewards in return. These platforms have gained immense popularity in recent years, with the incentives playing a pivotal role in attracting numerous funders [5]. Throughout the fundraising duration, campaigners utilize visuals, videos, narratives, endorsements, and pre-use testimonials to demonstrate their charity cause's authenticity within the campaign [6]. Moreover, they actively involve potential donors in online conversations and sometimes leverage donors as referrals to further amplify crowd engagement [7].

Charity crowdfunding has gained popularity in developing nations like India, where donations are incentivized through mechanisms like tax advantages [8]. Data from the Indian charity crowdfunding platform Ketto.org unveils a significant landscape: over 7,000 projects have garnered support from 320,000 backers, contributing a cumulative sum of 120 million (INR) by 2020 [9]. Established in 2012, Ketto.org stands as one of India's most prominent charity crowdfunding platforms. However, the ratio of charity campaigns achieving their initial fundraising targets is relatively low across various crowdfunding platforms, particularly in comparison to those backing startup projects instead of charitable causes [10]. For most charity fundraisers, the inability to gather sufficient funds within the stipulated time frame and in alignment with the original goal designates the campaign as a failure. In such cases, rerunning the campaign may become necessary if feasible [11]. This challenge significantly hampers the potential of charity fundraising endeavours to secure support for their causes [12]. This research centres on the assessment of influential attributes, specifically textual components, that may impact the success of fundraising for charity campaigns.

Prior research has predominantly approached crowdfunding platforms through an innovation and entrepreneurial lens, particularly focusing on campaigns aligned with technological startups. However, these investigations primarily scrutinized crowdfunding from a technological innovation standpoint and a concrete reward structure [13]. Additionally, numerous studies delved into the influence of quantitative attributes like funding amounts and project durations on crowdfunding achievement [14]. Some prior inquiries explored the impact of thematic attributes (latent semantics) extracted from textual campaign descriptions on their success [15,16]. Nonetheless, to date, no research has comprehensively explored the thematic elements that drive the success of charity fundraising campaigns and accurately predict their outcomes based on thematic attributes and numerical parameters [17]. This study effectively addresses these gaps in the existing literature.

In charity crowdfunding, individuals seeking funds, known as campaigners or beneficiaries, utilize crowdfunding websites to present their causes while providing incentives to donors [18]. The narratives outlining the projects and the incentives offered can serve as valuable sources for analysing and projecting the success of campaigns [19,20]. In this research endeavour, topical attributes were extracted as clusters of contextually associated terms (referred to as topics), such as medical needs, financial requirements, urgent expenses, and more. This extraction was achieved through the application of a topic modelling approach to the campaign descriptions featured on Ketto.org [21]. These topical attributes have proven effective in prognosticating the achievements of fundraising campaigns, as they align with people's inclination to support charitable causes across various donation tiers [22]. Within topic modelling, each topic encapsulates a concept with semantic coherence, reflecting real-world behaviours, actions, or conditions pertaining to charitable causes [19,23].

This research makes a significant contribution by combining topical attributes extracted from the dataset with standard numerical features to enhance the accuracy of predicting the success of crowdfunding campaign funding [24]. Unlike the traditional keyword-based approach, this study employs a topic-based text analytics method. In the topic-based approach, topics encompass sets of semantically coherent words, in contrast to the keyword-based method that assumes word independence [25]. Comparatively, the topic-based text analytics method is proven to be more effective than the keyword-based approach [26]. This study's contributions are threefold. Firstly, it introduces a novel text analytics framework for analysing and predicting the fundraising success of charity crowdfunding campaigns. Secondly, it employs a domain-constrained Latent Dirichlet Allocation (DC-LDA) topic model to enhance the extraction of topical features (latent semantics) from textual descriptions and narratives of charity campaigns to identify pertinent topics [27]. Lastly, the study conducts an empirical analysis to ascertain the features influencing campaign fundraising success.

This research marks the pioneering application of a topic modelling method to extract topical features from charity crowdfunding campaign content. From a managerial perspective, this study offers charity campaigners a tool to discern the most influential textual narratives impacting fundraising outcomes, aiding them in crafting time-sensitive campaigns around these narratives. The paper's structure is as follows: Section 2 provides an overview of prior literature concerning charity crowdfunding and identifies the research gap addressed in this study. Section 3 outlines the proposed text analytics methodology employed. Section 4 delves into the computational methodology utilized. Section 5 presents empirical findings and concludes with reflections and future research prospects.

## 2. LITERATURE REVIEW

### 2.1 Charity Crowdfunding

Originating from crowdsourcing, open call sourcing emerged as a means to comprehend diverse tasks using data generated within the social web [4,12]. Among its subtypes, crowdfunding stands out, encompassing fundraising endeavors for both new startup concepts and charitable causes. Functioning as an open call service, crowdfunding serves to acquire financial backing through donations (for charity) or rewards (for startups) [28]. A typical crowdfunding initiative involves a project initiator or lead campaigner, backers or donors (for charity), and crowdfunding platforms linking projects with donors [29,30]. Projects are categorized into non-profit or for-profit, falling under various types like equity-based, reward-based, loan-based, or donation-based. Existing research has primarily examined crowdfunding success factors from numerical aspects, lacking consideration of narrative text's perspective [31].

Prior research centered on comprehending the interplay between campaigners and backers, uncovering the dynamics driving successful campaign [32]. These studies revealed that campaigners gained insights into capital raising and honed funding skills through crowdfunding. Most backers were drawn in by rewards, support for novel concepts, and their contribution to societal causes. Limited investigations explored word-of-mouth effects and the sway of peer recommendations. In terms of forecasting crowdfunding triumph, existing literature primarily concentrated on project aspects like objectives, categories, rewards count, duration, multimedia content, and sharing channels [33]. These features were subsequently subjected to predictive analytics techniques to anticipate campaign success.

Different research has explored linguistic attributes extracted from campaign descriptions and numerical factors to anticipate crowdfunding triumph. Project quality concerning the idea and the impact of the campaigner's reputation have also been scrutinized for predictive insights [34]. Various machine learning techniques such as Support Vector Machine (SVM), Decision Trees (DT), K-nearest Neighbor (KNN), Markov Chain, and SVM have been employed [35,36]. While prior studies have extensively investigated predictive features in the startup crowdfunding realm, this study pioneers the examination of topical attributes and intangible incentive systems within charity-based crowdfunding.

## 2.2 Contribution of this Study

This study contributes in several distinctive ways. First, it delves into the literature surrounding charity crowdfunding, focusing notably on the realm of non-tangible incentives offered within these campaigns. This aspect has not been explored in previous literature, to our knowledge. Second, while prior research has applied topic modelling techniques to unveil topical attributes in tech startup campaigns, we uniquely apply this approach to charity crowdfunding platforms. The project descriptions on these platforms tend to be more specialized and tailored to sway donors toward the cause. Despite their non-tangible nature, incentives in charity campaigns bear similarities to reward-based incentives in startup endeavours. Here, we utilize topic modelling to dissect project descriptions and narratives that elucidate incentive information.

Thirdly, unlike previous studies that employed numerical parameters to predict the success of startup crowdfunding projects, our study adopts a fresh angle. We employ topical attributes and incentive-related attributes to predict the success of charity campaigns [37]. Lastly, while most prior studies predominantly scrutinized crowdfunding websites in developed countries, with a primary focus on startup crowdfunding, our research zeroes in on charity crowdfunding in India. This shift is significant as the motives and incentives are often aligned with community service and tax benefits.

## 3. ARTIFACT DESIGN

### 3.1 Text Analytics Framework for Charity Crowdfunding Analysis

Project descriptions exert a substantial impact on campaign outcomes, and earlier research has predominantly approached this subject using numerical indicators such as word count and grammatical accuracy to forecast success [7]. The assessment of shallow linguistic aspects has been a central facet in gauging campaign efficacy. The preceding literature predominantly revolved around the utilization of numerical parameters to determine campaign triumph, employing them as yardsticks for success and comparative analysis among projects [38]. A limited subset of studies has ventured into exploring topical characteristics as determinants of campaign success, with an even smaller fraction delving into the role of rewards as antecedents [14]. These investigations have primarily centered on startup crowdfunding, spanning both reward-based and non-reward-based models [39]. However, a gap exists in focusing on charity crowdfunding as a domain for the examination of topical attributes and their predictive potential in influencing campaign outcomes [40].

This study enhances the traditional Latent Dirichlet Allocation (LDA) model by implementing it within the context of charity crowdfunding. This novel model is termed the Domain-Specific LDA (DS LDA), demonstrating improved topic modeling capabilities [41]. The topical attributes generated through DS LDA are subsequently harnessed within a random forest classifier for campaign success prediction [42]. The devised text analytics framework for charity crowdfunding analysis is illustrated in Fig. 1, comprising three tiers: data collection, feature extraction, and prediction.

### 3.2 Data Collection

Data for the project was amassed from Ketto.org, a renowned charity crowdfunding platform in India. This platform has earned repute for its effectiveness in supporting charitable causes across diverse categories, resulting in a notable campaign success rate. Since its establishment in 2012, Ketto.org has fostered a donor network exceeding 5.5 million individuals, contributing approximately 300 million USD in donations. The platform boasts user-friendly navigation and empowers users to craft their personalized online fundraising campaign pages, facilitating fundraising for charitable and individual initiatives.

The acquired data encompasses textual elements, such as project descriptions featuring the extracted topical attributes, and numerical components including the goal amount, raised amount, start and end dates, days remaining, top donor contribution, minimum donor contribution, and supporter count. The specifics of these numerical attributes are elaborated in Table 1. This set of seven attributes is employed within the proposed predictive framework, forming the basis for constructing the random forest model aimed at forecasting campaign success.

Regarding the textual attributes, both campaign descriptions and incentive descriptions are incorporated. These two forms of descriptions exhibit distinct semantics, with campaign descriptions primarily focused on delineating the cause's nature and narrative, whereas incentive descriptions elucidate the rewards extended to donors.

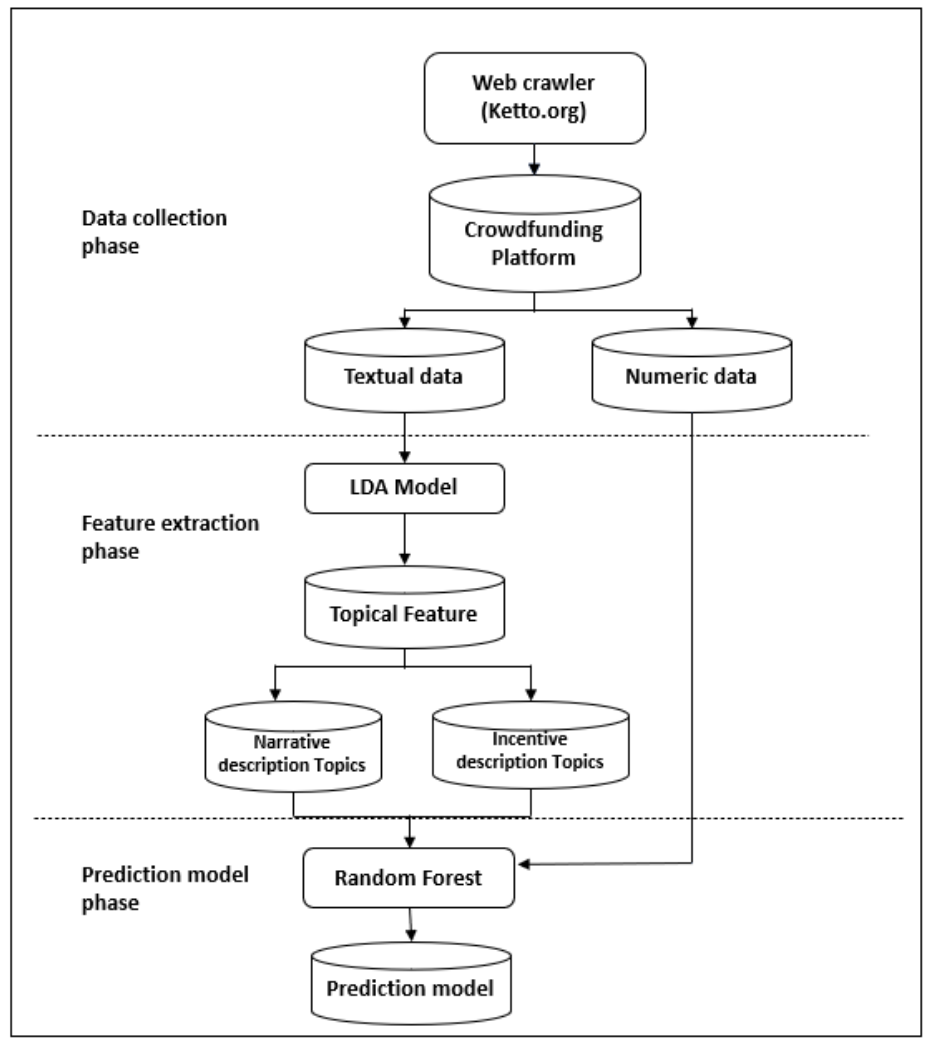

**Fig. 1. LDA Semantic text analytics artifact**

**Table 1. Numerical project features**

| Feature | Description |
|---|---|
| Goal amount | Targeted funding amount |
| Raised amount | Current/ final raised amount |
| Start date | Starting date of campaign/project |
| End date | End date of campaign/ project |
| Days left | Number of days left for campaign/ project closing |
| Top donor amount | Maximum donation donated |
| Minimum donor amount | Minimum donation donated |
| Number of supporters | Total number of donors |

### 3.3 Feature Extraction

Feature extraction via classification analysis proves pivotal as it classifies texts into topics, rendering it a crucial procedure. Following web data extraction, sentences undergo word segmentation, disassembling them into words and determining each word's part-of-speech (POS). In this analysis, solely nouns are deemed significant for extraction, being the most illustrative document tokens. To optimize noun extraction, stop words are excluded. The resulting topical attributes and incentives are then integrated with the seven enumerated numerical attributes from Table 1 to anticipate campaign success.

### 3.4 Prediction

For prediction purposes, a Random Forest classifier is employed to analyze and interpret the data. Random Forest, rooted in decision tree methodology, is harnessed to forecast campaign success. The data is transformed, with each campaign assigned a label of "1" if the funding goal is attained, or "0" if not. The dataset is then partitioned into training and test sets. The training set is employed to educate the Random Forest classifier using the dataset. This phase entails training separate decision trees using random subsets of the training set. The training process culminates in the construction of a predictive model based on the Random Forest, aiming to forecast campaign success. Ultimately, the predictors, encompassing both textual and numerical attributes, are scrutinized for their notable impact on the dependent variable of campaign success.

### 4. RESULTS

### 4.1 The Dataset

The dataset for crowdfunding was obtained through web crawling on the charity crowdfunding platform Ketto.org, utilizing a personally developed Python crawler rather than automated web tools. The crawler effectively harvested data from 410 projects categorized as medical initiatives among various project categories. The medical category was chosen due to its inherent urgency and tight deadlines. This choice was intended to minimize selection bias, incorporating completed projects that could potentially be reopened if their funding goals weren't met. Medical emergencies demand swift financial support to fulfil their obligations. In this study, out of the 410 projects, 210 achieved successes while 200 were unsuccessful. For prediction purposes, the seven numerical parameters were combined with the derived topical attributes to forecast campaign success. The dataset was split into a training set comprising 250 projects and a test set containing 160 projects to maintain a balanced distribution across classes.

### 4.2 Performance Measures

The proposed text analytics framework's effectiveness was assessed using performance measures. Four specific performance measures were employed to evaluate the prediction task's effectiveness. These measures are presented in Table 2.

The data from the crowdfunding platform corpus was subjected to the DS LDA model. The Gibbs sampler was utilized to approximate the topic-word distribution and the document-topic distribution within the text corpus. The data encompassed two document subsets: project descriptions containing topic words from cause narratives and incentive descriptions encompassing rewards, whether tangible or non-tangible. The empirical analysis determined the appropriate number of topics to characterize each text corpus based on the extracted words.

Perplexity measures were accurately computed for each textual corpus segment, be it cause narration or incentive narration. These perplexity scores were associated with topics within each textual corpus, and the topics were labelled based on their words' genre and semantic similarity. The analysis successfully ascertained a specific number of topics from the textual corpus, identifying variables that significantly enhanced the predictive capability of a model.

### 4.3 Topical Feature Selection

The proposed text analytics framework, coupled with the extended LDA model DS LDA, facilitates the extraction of topical attributes from both project descriptions and incentive descriptions, thereby aiding the assessment of charity crowdfunding campaign success. The DS LDA model adeptly mines topical features rooted in the intrinsic attributes of crowdfunding projects, drawing insights from the project narratives and descriptions. Moreover, the DS LDA model applies topic modelling to unravel the distinct incentives proffered to donors. The textual data related to these incentives is isolated into a separate corpus for analysis.

Initially, the topical features explored exhibited redundancy, complicating examination. To address this concern, a feature selection technique was employed to prune redundant terms from the corpus. The resulting extracted topical attributes from the project descriptions of the Ketto.org website is detailed in the table.

Derived from the textual corpus, the DS LDA successfully extracted two topics, as depicted in Table 3. Topic 1 centred on the medical needs of children, particularly those under 12 afflicted by terminal ailments. Significantly elevated attention was accorded to cases featuring young beneficiaries grappling primarily with terminal illnesses. Additionally, subcategories encompassed cases of children with serious injuries from accidents, chronic yet non-terminal illnesses, and minor ailments necessitating consistent care. In contrast, Topic 2 revolved around urgent care for the elderly, linked to campaigns necessitating immediate funding for closure. The majority of these campaigns exhibited deadlines spanning 48 to 72 hours, with fundraising goals ranging from $4,000 to $20,000. Notably, the most prosperous campaigns often featured lower goal amounts coupled with extended deadlines. Successful campaigns pertained to immediate surgeries and hospitalizations required for the elderly. Similarly, campaigns centred on sustaining ongoing treatments for the elderly achieved success.

These two topics held significant prominence within the textual features pertaining to campaign narration and project description. While this study did not incorporate sentiment analysis to assess the emotions and sentiments of donors, it was evident from the campaign success rates that endeavours featuring children in medical emergencies were notably successful. Extending this research to incorporate an analysis of donor sentiments and emotions towards different beneficiary types and their medical needs would provide intriguing insights for future investigations.

Table 4 illustrates two distinct topic categories explored within incentive descriptions. The first topic, Topic 1, predominantly encompasses terms related to tax reduction, tax benefits, and tax cuts. Tax reduction incentives are held in high regard within communities where charitable endeavours leverage government policies to offer benefits via tax exemptions. Many of these campaigns pertain to nonprofit organizations championing enduring causes such as poverty alleviation, economic needs, and societal improvements.

Meanwhile, Topic 2 concentrates on non-tangible incentives like website acknowledgments, certificates, honorary mentions in brochures, and award naming. Both incentive topics are prominently linked to extended-duration campaigns administered by nonprofit organizations.

## 4.4 Predictive Model

To assess the prediction capabilities of the proposed Random Forest classifier, its performance was contrasted with alternative classifier techniques including SVM, back-propagation neural network (BPNN), and extreme learning machine (ELM). The numerical feature set was sourced from Table 1, combined with the topical attributes derived using DS LDA from both campaign and incentive descriptions. In terms of accuracy, precision, recall, and F1 score, the Random Forest classifier outperformed the other classifiers, affirming its superior predictive efficacy.

### Table 2. ML Performance Metrics [1]

| ML Metrics | Definition | Formula |
|---|---|---|
| Accuracy | The ratio of number of correct predictions to the total number of predictions | $\frac{(TP+TN)}{(TP+TN+FP+FN)}$ |
| Precision | The ratio of number of correctly predicted positive values to the total predicted positive values | $\frac{TP}{(TP+FP)}$ |
| Recall | The ratio of correctly predicted positive values to the total number of positive values | $\frac{TP}{(TP+FN)}$ |
| F1 score | Harmonic mean of precision and recall | $\frac{2*(Precision*Recall)}{(Precision + Recall)}$ |

**Table 3. Representative Topical Features (Campaign descriptions)**

| Topic | Words |
|---|---|
| 1 | Child health, cancer, terminal disease, terminal cancer, leukemia, central nervous system (CNS) tumors, lymphomas, severe illness, parent dependence |
| 2 | Elderly health, nursing facility, kidney dialysis, diabetes, heart stroke, urgent funds, urgent hospitalization, arthritis, physiotherapy |

**Table 4. Representative Topical Features (Incentive descriptions)**

| Topic | Words |
|---|---|
| 1 | Tax benefit, tax reduction, income tax benefits, section 80G, medical relief, section 12AA |
| 2 | Appreciation post, certificate, certificate of donation, recognition, voluntary board member, request to be guest, induction as honorary |

**Table 5. Prediction performance of various classifiers**

| Measures | Random Forest | SVM | BPNN | ELM |
|---|---|---|---|---|
| Accuracy | 78.00 | 56.00 | 48.00 | 42.00 |
| Precision | 72.42 | 54.26 | 46.88 | 48.00 |
| Recall | 88.66 | 82.22 | 84.44 | 78.76 |
| F1 score | 82.56 | 78.23 | 74.72 | 66.32 |

## 5. CONCLUSION AND FUTURE WORK

Charity crowdfunding platforms have gained popularity, yet academic research in this domain remains largely unexplored. While earlier studies have predominantly concentrated on dissecting topical elements and predictors that contribute to the success of startup crowdfunding ventures, this study takes a novel stride by investigating the topical attributes and predictors that potentially drive success in charity campaigns. Focusing on medical-based needs, which are characterized by tight deadlines and constrained resources, adds a time-sensitive context to the research. The findings of this study underscore the dominance of two prevailing topics associated with beneficiary types: medical requirements for children and the elderly. Furthermore, the scope of incentive-related topics is narrowed down to tangible rewards, such as tax benefits, and intangible ones, like recognition. Overall, this study paves the way for further exploration within the realm of charity crowdfunding. Future research avenues include incorporating sentiment analysis to grasp donor sentiments and emotions towards particular beneficiaries or charitable causes. Additionally, expanding this research to encompass developed countries and diverse incentive types could yield valuable insights. Lastly, delving into donor comments can provide a deeper understanding of the emotions and sentiments stirred by the causes among donors.

## 6. PRATICAL IMPLICATION

This research study carries two practical implications. Firstly, it provides guidance for charity funding institutions and networking sites, assisting them in optimizing their budget and efforts by emphasizing factors that have a greater appeal to potential donors. Secondly, for academic researchers, it contributes to a deeper understanding and development of IT artifacts, showcasing innovative text analytics and predictive models. In essence, this study serves as a foundational step in establishing analytical modelling within the domain of crowd charity funding, offering valuable insights for both practitioners and scholars alike.

## COMPETING INTERESTS

Authors have declared that no competing interests exist.